\documentclass[10pt,twocolumn,letterpaper]{article}

\usepackage[pagenumbers]{cvpr} % To force page numbers, e.g. for an arXiv version

\usepackage{tikz}
\usepackage{amsmath,amssymb,amsthm}
\usepackage{graphicx}
\usepackage{booktabs}

\usepackage[table]{xcolor}
\usepackage{booktabs}
\usepackage{amssymb} % for up/down arrows
\newcommand{\best}[1]{\textbf{#1}}
\newcommand{\second}[1]{\underline{#1}}

\definecolor{cvprblue}{rgb}{0.21,0.49,0.74}
\usepackage[pagebackref,breaklinks,colorlinks,allcolors=cvprblue]{hyperref}
\usepackage{soul}
\usepackage{xcolor}

\def\paperID{*****} % *** Enter the Paper ID here
\def\confName{CVPR}
\def\confYear{2026}

\title{Dissecting Model Failures in Abdominal Aortic Aneurysm Segmentation through Explainability-Driven Analysis}

\author{
Abu Noman Md Sakib$^{1,*}$ \and
Merjulah Roby$^{1,*}$ \and
Zijie Zhang$^{1}$ \and
Satish Muluk$^{2}$ \and
Mark K. Eskandari$^{3}$ \and
Ender A. Finol$^{1}$ \and
{\small$^{1}$University of Texas at San Antonio $^{2}$Drexel University $^{3}$Northwestern University}
}

\begin{document}
\maketitle
\begingroup
\renewcommand{\thefootnote}{\fnsymbol{footnote}}
\footnotetext[1]{These authors contributed equally to this work.}
\endgroup
\setcounter{footnote}{0}
\begin{abstract}
Computed tomography image segmentation of complex abdominal aortic aneurysms (AAA) often fails because the models assign internal focus to irrelevant structures or do not focus on thin, low-contrast targets. Where the model looks is the primary training signal, and thus we propose an Explainable AI (XAI) guided encoder shaping framework. Our method computes a dense, attribution-based encoder focus map ("XAI field") from the final encoder block and uses it in two complementary ways: (i) we align the predicted probability mass to the XAI field to promote agreement between focus and output; and (ii) we route the field into a lightweight refinement pathway and a confidence prior that modulates logits at inference, suppressing distractors while preserving subtle structures. The objective terms serve only as control signals; the contribution is the integration of attribution guidance into representation and decoding. We evaluate clinically validated challenging cases curated for failure-prone scenarios. Compared to a base SAM setup, our implementation yields substantial improvements. The observed gains suggest that explicitly optimizing encoder focus via XAI guidance is a practical and effective principle for reliable segmentation in complex scenarios.
\end{abstract} 

\section{Introduction}
Abdominal aortic aneurysm (AAA) is a life-threatening vascular disease in which local dilation of the abdominal aorta can progress silently and culminate in catastrophic rupture. Accurate segmentation of AAA structures from medical images is critical for risk assessment, surgical planning, and longitudinal monitoring \cite{aaamortality, aaamortality2}. In clinical practice and research pipelines, two masks are typically of interest: the outer aneurysm wall and the lumen \cite{aaaautomatic}. Although modern deep segmentation architectures including U-Net variants \cite{unet, unetpp, unetpp2, li2025reseg} and, more recently, Segment Anything Model (SAM) style large vision encoders have achieved strong performance on relatively clean cases, their predictions often break down in the complex, failure-prone scenarios that are most clinically consequential \cite{sam, medsam, ke2023segment, ren2024segment, mazurowski2023segment, tai2025segment}.

\begin{figure}[t]
    \centering
    \includegraphics[width=\columnwidth]{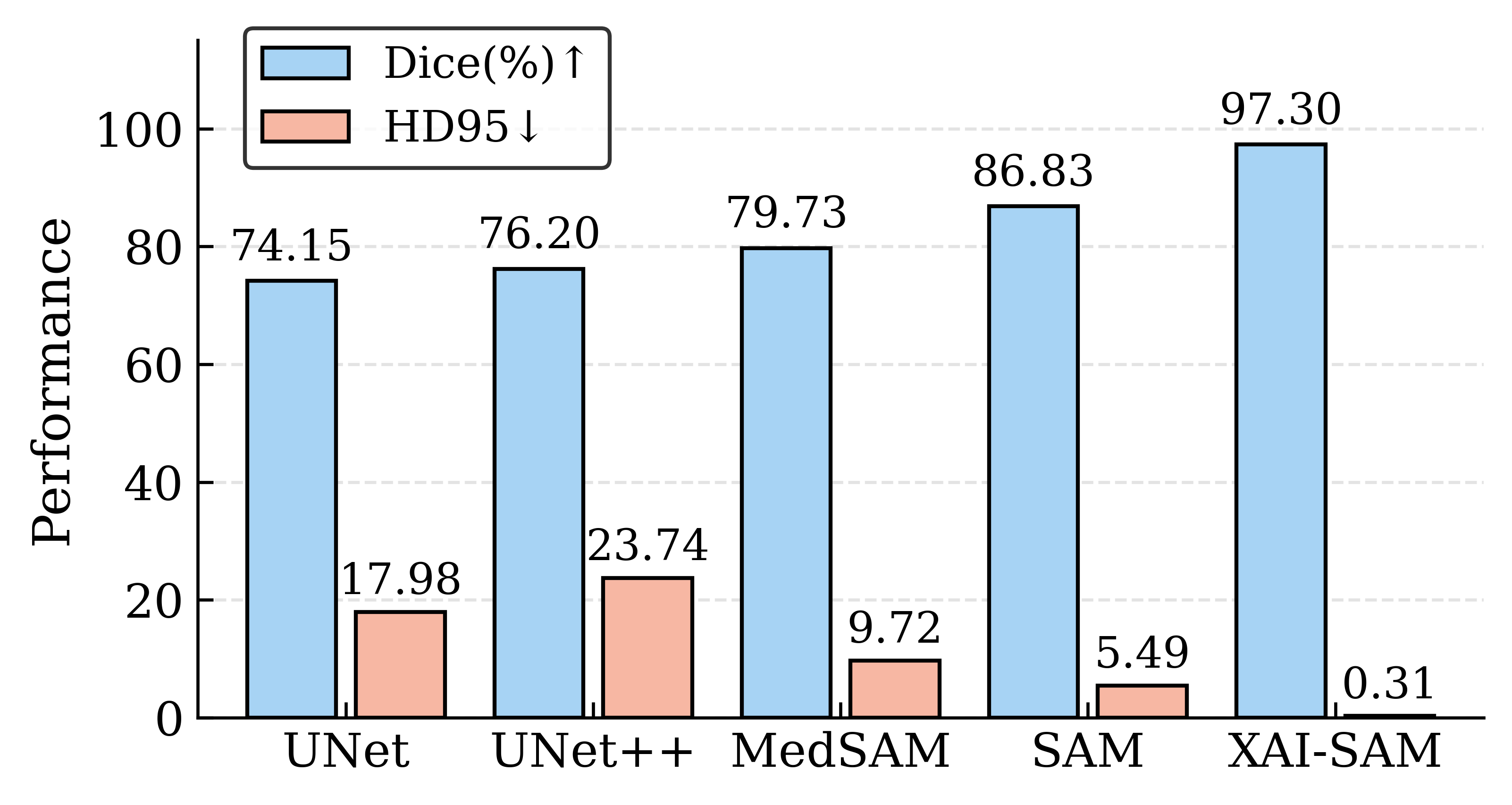}
    \caption{Performance comparison in complex outer wall AAA segmentation cases. Our XAI-SAM method significantly outperforms U-Net, U-Net++, MedSAM, and SAM measured by Dice score and HD95.}
    \label{fig:comparison}
\end{figure}

In challenging AAA cases, segmenters frequently allocate the probability mass to the wrong anatomical structures (e.g., nearby vessels or soft tissue) or leak into regions that should not be segmented. These errors are exacerbated when the aneurysm wall is thin, has low contrast against the surrounding tissue, or exhibits irregular morphology. Conventional training objectives such as Dice loss or cross-entropy treat these failures implicitly through pixel-wise errors, but they do not directly reason about where the model is focusing internally when it makes these mistakes. As a result, the learned encoder representation may remain misaligned with the clinically relevant structures even when the mask losses are minimized.

In this work, we treat the internal focus of the encoder as the primary objective of optimization rather than as a by-product. We propose an explainability-guided framework that uses attribution maps derived from the final encoder block as an explicit training signal. We compute a dense, gradient-based attribution field, which we term an Explainable AI (XAI) field, and integrate it into both representation learning and decoding. First, we encourage alignment between the predicted segmentation probability mass and the XAI field. This strategy highlights the agreement between where the model looks and where it predicts. Second, we route the XAI field into a lightweight refinement pathway and a confidence prior that modulate logits, suppressing distractors while preserving subtle structures. In parallel, we introduce a pairwise consistency regularizer based on a classifier trained on consecutive mask pairs, which discourages anatomically implausible transitions across slices.

We evaluated our method on a clinically curated AAA dataset that explicitly includes complex, failure-prone cases for both outer wall and lumen segmentation. Across the general test set, our approach substantially outperforms strong baselines, including U-Net \cite{unet}, U-Net++ \cite{unetpp}, MedSAM \cite{medsam}, and a SAM-based \cite{sam} segmentation baseline. In the subset of complex cases, the gains are even greater, with our \textbf{XAI}-guided \textbf{SAM} model (XAI-SAM) closing the majority of the failure modes observed in the baselines (Fig. \ref{fig:comparison}). Beyond aggregate metrics, qualitative case studies show that our method refines internal focus away from irrelevant structures and towards thin aneurysm walls and lumen boundaries. Our main contributions are to
\begin{itemize}
    \item Propose an XAI-guided encoder shaping framework that treats encoder attribution maps as a first-class training signal for AAA segmentation.
    \item Introduce a focus alignment loss that enforces agreement between segmentation probability mass and an attribution-based XAI field, and we couple this with a refinement pathway and confidence prior that modulate logits.
    \item Incorporate a pairwise consistency classifier trained in consecutive ground-truth masks and use it as a regularizer to penalize anatomically inconsistent predictions across slices.
    \item Conduct a detailed empirical study of outer wall and lumen masks in both general and complex AAA cases, which shows consistent and substantial improvements over U-Net, U-Net++, MedSAM, and SAM baselines, accompanied by qualitative and explainability-driven analyses of model failures and corrections.
\end{itemize}

\section{Related Work}

\begin{figure*}[t]
    \centering
    \includegraphics[width=\textwidth]{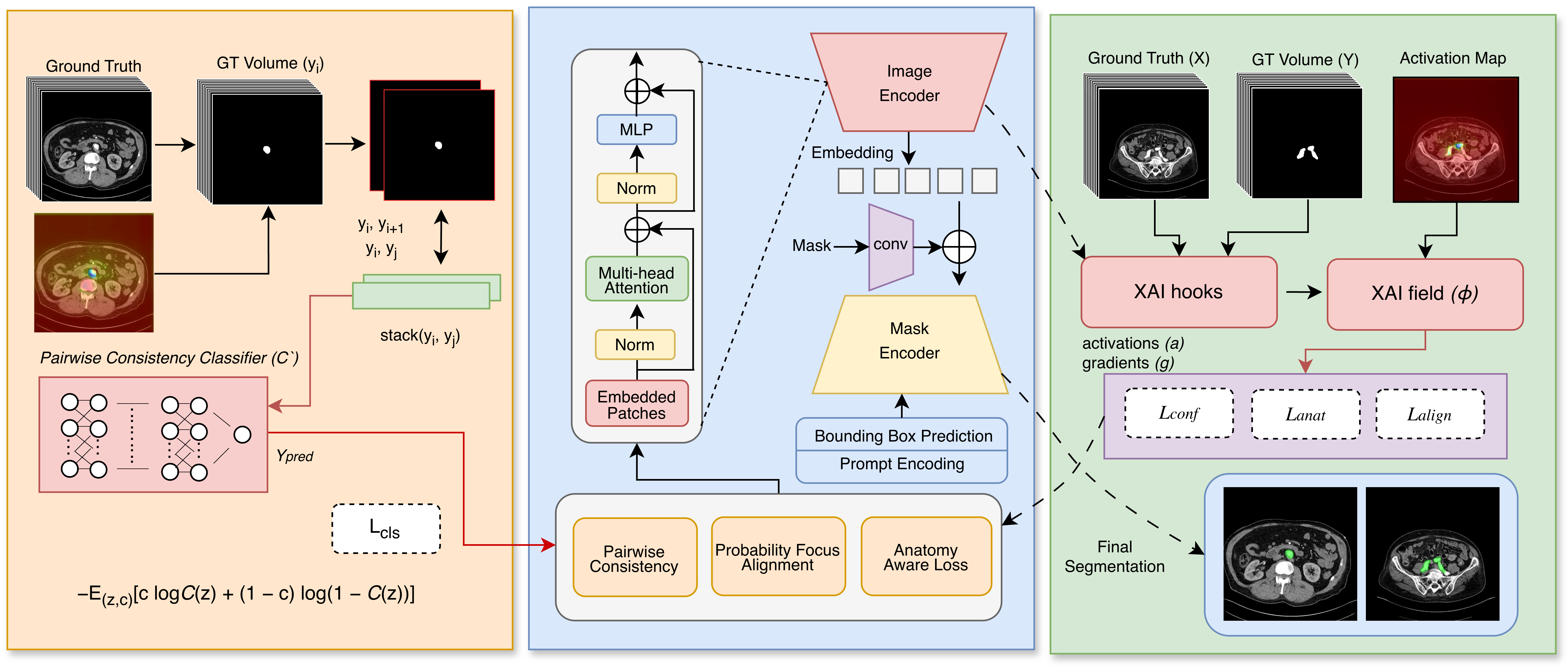}
    \caption{Overview of the proposed XAI-SAM framework. The left module introduces the pairwise consistency mechanism. The central module shows the SAM-based segmentation pipeline enhanced with pairwise consistency, probability-focus alignment, and anatomy-aware loss. The right module illustrates the XAI component.}
    \label{fig:methodology}
\end{figure*}

\subsection{AAA Segmentation and Vascular Imaging}
Automated segmentation of AAA structures has been explored using classical image processing pipelines, active contour models, and, more recently, convolutional neural networks \cite{lyu2024automatic, wan2024bif}. Early approaches often relied on intensity thresholds, region growing, or hand-crafted features tailored to contrast-enhanced CT or MR data. These methods are sensitive to noise, calcifications, and irregular aneurysm morphology, and typically struggle in difficult cases \cite{mu2023automatic}. With the advent of deep learning, U-Net and its variants have become the standard for medical image segmentation, including vascular and aneurysm tasks \cite{rainville2025weakly}. However, even these architectures can fail on thin, low-contrast aneurysm walls or in the presence of adjacent vessels with similar intensity profiles.

\subsection{Deep Segmentation Architectures}
Encoder–decoder architectures such as U-Net, U-Net++, and related designs have achieved state-of-the-art performance in many medical segmentation benchmarks by combining multi-scale feature extraction with skip connections \cite{vaswani2017attention, unet, unetpp, ding2025mg, cao2022swin, isensee2021nnu}. Extensions incorporate attention modules, residual blocks, or multi-branch decoders to better capture context and fine structures \cite{ibtehaz2023acc, xu2024lb, ruan2023ege}. Although self-configuring frameworks such as nnU-Net provide a robust benchmark for medical tasks, they rely primarily on standard volumetric losses \cite{isensee2021nnu}. In parallel, large-scale vision models, such as the Segment Anything Model (SAM), have demonstrated strong zero-shot and prompt-driven segmentation abilities in natural images \cite{sam, wang2024segment, kweon2024sam, li2024segment}. MedSAM-style adaptations have begun to transfer this capability to medical imaging domains, providing powerful encoders and flexible prompt-aware decoders \cite{medsam, pandey2023comprehensive}. Despite their capacity, these models are typically optimized using standard region-based losses and are not explicitly encouraged to focus on clinically relevant structures. In failure-prone AAA cases, they may assign high probability to distractor regions even when global metrics remain acceptable on easier slices.

\subsection{Explainability and XAI-guided Learning}
Explainability methods such as Grad-CAM \cite{selvaraju2017grad, chattopadhay2018grad}, integrated gradients \cite{kapishnikov2021guided}, and related attribution techniques have been widely used to visualize how models make decisions \cite{wang2020score}. In segmentation, attribution maps can reveal whether the internal focus of a model aligns with the target structure or is driven by spurious cues. Although most work uses explainability for post-hoc analysis, recent studies have begun to incorporate attribution into training objectives, for example by penalizing attention on known confounders or encouraging focus on salient regions \cite{seggradcam, bach2015pixel, sundararajan2017axiomatic, smilkov2017smoothgrad}. However, such approaches are still relatively rare in medical imaging, and few works integrate attribution signals directly into the encoder representation and decoder logits in a unified way.

\subsection{Shape and Consistency Constraints}
Several works have explored shape priors, topology-aware losses, and consistency constraints to encourage anatomically plausible segmentation. Skeleton-based losses (e.g., clDice) and edge-aware losses (e.g., Sobel-based boundary Dice) have been proposed to better preserve thin structures and connectivity \cite{chen2023differentially, mao2024realizable}. Temporal or slice-wise consistency has also been studied, where predictions across neighboring frames or slices are regularized to align with motion or anatomical continuity \cite{mao2023cross}. Our method is inspired by this line of work but takes a different route: instead of handcrafting pairwise penalties, we train a small classifier on pairs of ground-truth masks to learn what ``realistic'' slice-to-slice consistency looks like, and then use this classifier as a learned regularizer during segmentation training.

\section{Methods}

\subsection{Problem Setup and Backbone}
Let $x \in \mathbb{R}^{3 \times H_0 \times W_0}$ denote an input AAA image slice, rescaled to $H_0 = W_0 = 1024$. Each slice has two binary masks: one for the outer aneurysm wall and one for the lumen. For simplicity, we describe the method for a single binary target $y \in \{0,1\}^{1 \times H \times W}$. The base architecture is based on an SAM-style encoder–decoder. Although the framework is backbone-agnostic, we focus on SAM-based models due to their tendency toward diffuse attention in medical images. A shallow pre-encoder $M$ maps the input to a normalized representation, $x' = M(x) \in \mathbb{R}^{3 \times H \times W}$.

A Vision Transformer (ViT)-based \cite{yin2022vit} image encoder $\mathrm{Enc}$ processes the normalized input $x'$ and produces feature maps $E = \mathrm{Enc}(x') \in \mathbb{R}^{C \times H_e \times W_e}$, where $C = 256$ represents the number of channels. These feature maps are further refined by applying a channel-wise MLP with residual connections: $E' = F(E) = E + \mathrm{MLP}(E)$. Next, we employ a projection head $P$ to predict a bounding box prompt $b = P(E) \in \mathbb{R}^4$, which is subsequently fed into the SAM prompt encoder and mask decoder to generate segmentation logits. The logits are then passed through a sigmoid function, yielding segmentation probabilities $p = \sigma(s)$, where $s = \mathrm{Dec}(E, b) \in \mathbb{R}^{1 \times H \times W}$ represents the final segmentation logits.

In parallel, an auxiliary decoder $G$ is attached to the shaped features $E'$, producing auxiliary probabilities $a = G(E') \in \mathbb{R}^{1 \times H \times W}$. These auxiliary probabilities are passed through a sigmoid activation to obtain $p_{\mathrm{aux}} = \sigma(a)$. A confidence head $A$ takes $p_{\mathrm{aux}}$ as input and maps it to a refinement prior $m_c = A(p_{\mathrm{aux}}) \in [0,1]^{1 \times H \times W}$. In inference, this confidence prior is used to modulate the final logits, thereby suppressing distractor regions while reinforcing confident predictions. Fig. \ref{fig:methodology} provides a detailed illustration of our architecture, integrating pairwise consistency, anatomy-aware learning, and explainability-guided optimization to enhance both segmentation accuracy and interpretability.

\subsection{Failure Analysis of Baseline}
\label{sec:failure-analysis}

\begin{figure}[htbp]
    \centering
    \includegraphics[width=\columnwidth]{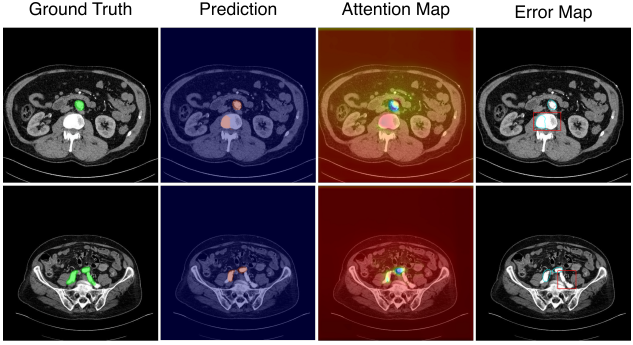}
    \caption{Visualization of model performance on abdominal CT slices. From left to right: Ground Truth segmentation, Model Prediction, Attention Map highlighting regions of model focus, and Error Map indicating mismatched regions (red box) between Prediction and Ground Truth.}
    \label{fig:xai-prob-analysis}
\end{figure}

Despite strong aggregate scores on easy slices, SAM-style baselines exhibit systematic errors on AAA: (\textit{i}) probability mass concentrates on distractors (adjacent vessels, calcifications), (\textit{ii}) over-segmentation into background or lumen, and (\textit{iii}) abrupt, anatomically implausible changes across consecutive slices (Fig. \ref{fig:xai-prob-analysis}). We hypothesize that these behaviors correlate with encoder focus misalignment, i.e., internal attribution peaks do not coincide with clinically relevant structures. We therefore instrument the encoder with explainability probes and derive quantitative failure indices. 

For the attribution probe, let $\phi \in \mathbb{R}^{1 \times H \times W}$ denote a gradient-weighted attribution map obtained from the last encoder block (as in \S\ref{sec:xai-field}, using activations $\mathcal{A}$ and gradients $\mathcal{G}$ with channel weights $w_c = (H_eW_e)^{-1}\sum_{i,j}\mathcal{G}_{c,i,j}$ and $\phi_{\mathrm{low}}=\mathrm{ReLU}(\sum_c w_c \mathcal{A}_c)$, upsampled to $(H,W)$ and $\ell_1$-normalized to $\tilde{\phi}$). Let the baseline posterior be $p=\sigma(s)\in[0,1]^{H\times W}$ and the ground-truth mask be $y\in\{0,1\}^{H\times W}$. We measure how the distribution of predicted probability $\tilde{p} \doteq p/\sum p$ deviates from the normalized attribution $\tilde{\phi}$ using a symmetrized KL divergence:
\begin{equation}
\mathrm{JSD}(\tilde{p},\tilde{\phi}) \;=\; \tfrac{1}{2}\,\mathrm{KL}\!\left(\tilde{p}\middle\|\tfrac{\tilde{p}+\tilde{\phi}}{2}\right) \;+\; \tfrac{1}{2}\,\mathrm{KL}\!\left(\tilde{\phi}\middle\|\tfrac{\tilde{p}+\tilde{\phi}}{2}\right)
\label{eq:jsd}
\end{equation}
Large $\mathrm{JSD}$ indicates scattered or off-target focus even when region losses are small. We define a focus overlap index and its complement:
\begin{equation}
\mathrm{FOI} \;=\; \frac{\sum_{i,j}\phi_{i,j}\,y_{i,j}}{\sum_{i,j}\phi_{i,j}+ \epsilon},
\qquad
\mathrm{FMI} \;=\; 1-\mathrm{FOI}
\label{eq:fmi}
\end{equation}
Here, $\mathrm{FMI}$ signals that the encoder attends outside the aneurysm region. To expose over-segmentation, we quantify attribution mass on the complement of the mask:
\begin{equation}
\mathrm{Leak}_{\phi} \;=\; \frac{\sum \phi\,(1-y)}{\sum \phi+\epsilon},
\qquad
\mathrm{Leak}_{p} \;=\; \frac{\sum p\,(1-y)}{\sum p+\epsilon}
\label{eq:leak}
\end{equation}
To test sensitivity to thin walls, let $\partial y$ be a (one-pixel) morphological boundary of $y$ and $B_r(\partial y)$ its $r$-dilation; the \emph{boundary coverage} is
\begin{equation}
\mathrm{BCov}_r \;=\; \frac{\sum_{(i,j)\in B_r(\partial y)} \phi_{i,j}}{\sum \phi + \epsilon},
\label{eq:bcov}
\end{equation}
with low $\mathrm{BCov}_r$ indicating failure to concentrate on the true wall boundary. Given consecutive predictions $\hat{y}_i=\mathbb{1}[p_i>0.5]$ and $\hat{y}_{i+1}$, we compute a boundary-based chamfer distance $d_\mathrm{ch}(\partial\hat{y}_i,\partial\hat{y}_{i+1})$ and define
\begin{equation}
\mathcal{E}_{\mathrm{cons}} \;=\; \frac{1}{|\mathcal{N}|}\sum_{i\in\mathcal{N}} d_\mathrm{ch}(\partial\hat{y}_i,\partial\hat{y}_{i+1})
\label{eq:consistency}
\end{equation}
where $\mathcal{N}$ indexes valid slice pairs. A high $\mathcal{E}_{\mathrm{cons}}$ reflects anatomically implausible jumps along the stack. In failure-prone slices, we observe (\textit{i}) elevated $\mathrm{JSD}(\tilde{p},\tilde{\phi})$ and $\mathrm{FMI}$, (\textit{ii}) high $\mathrm{Leak}_{\phi}$/$\mathrm{Leak}_{p}$ together with low $\mathrm{BCov}_r$, and (\textit{iii}) spikes in $\mathcal{E}_{\mathrm{cons}}$. In addition, the Spearman correlations $\rho(\mathrm{JSD},\,1{-}\mathrm{IoU})$ and $\rho(\mathrm{FMI},\,1{-}\mathrm{Dice})$ are strongly positive, indicating that focus misalignment predicts segmentation error. Visual overlays of $\phi$ on the baseline SAM predictions confirm attention concentrated on distractors precisely where IoU collapses.

This analysis motivates two principles substantiated in XAI-SAM: (i) focus alignment: explicitly shape the encoder so that predicted probability mass agrees with attribution (reducing $\mathrm{JSD}$ \eqref{eq:jsd}, and $\mathrm{FMI}$ \eqref{eq:fmi}); and (ii) anatomy and temporal regularization: bias focus toward boundary neighborhoods (raising $\mathrm{BCov}_r$ \eqref{eq:bcov}) and suppress over-segmentation (lower $\mathrm{Leak}_{\phi}$/$\mathrm{Leak}_{p}$ \eqref{eq:leak}), while enforcing slice-to-slice plausibility (lower $\mathcal{E}_{\mathrm{cons}}$ \eqref{eq:consistency}). The subsequent subsections operationalize these principles via our alignment losses, confidence prior, and learned regularizer.

\subsection{Pairwise Consistency Classifier}
We first address failures where the model segments the wrong region or exhibits inconsistent masks across consecutive slices. To mitigate failures, we introduce a learned slice–pair consistency module. The idea is to constrain predicted masks to lie on a manifold of realistic anatomical transitions observed in ground-truth 3D volumes.

Given a stack of binary masks $\{y_i\}_{i=1}^N$ for a training volume, we construct positive examples using consecutive pairs $(y_i, y_{i+1})$ and negative examples by sampling non-consecutive pairs $(y_i, y_j)$ with $|i-j|>1$. Each pair is represented as a two-channel tensor $z=\mathrm{stack}(y_i,y_j)\in\{0,1\}^{2\times H\times W}$ with an associated binary label $c\in\{0,1\}$ indicating whether it forms a true anatomical progression. A lightweight CNN classifier $C$ maps $z$ to a consistency score $C(z)\in(0,1)$ and is trained using the binary cross-entropy objective 
\begin{equation}
\mathcal{L}_{\mathrm{cls}}
= -\mathbb{E}_{(z,c)}\!\left[c\log C(z) + (1-c)\log(1-C(z))\right]    
\end{equation}

During segmentation training, the classifier acts as a learned anatomical prior. For each minibatch $\{(x_i,y_i)\}_{i=1}^B$, predicted masks are first computed as $p_i=\sigma(s_i)$ and then binarized as $\hat{y}_i = \mathbb{1}[p_i>0.5]$. Forward and backward inter-slice penalties are evaluated by feeding pairs such as $(\hat{y}_i, y_{i+1})$ or $(y_{i-1},\hat{y}_i)$ through $C(\cdot)$ and accumulating penalties of the form $1 - C(\cdot)$. If $\ell$ denotes any such per-pair penalty and $N_{\mathrm{pairs}}$ the number of valid neighboring slice pairs in the minibatch, we compute the slice-consistency energy simply as the average penalty, i.e., $\mathcal{L}_{\mathrm{cons}} = N_{\mathrm{pairs}}^{-1}\sum \ell$. This quantity is then scaled by a weighting coefficient $\lambda_c$ to obtain the final consistency regularizer, expressed compactly as $\mathcal{L}_{\mathrm{pair}} = \lambda_c\,\mathcal{L}_{\mathrm{cons}}$.

If the classifier $C$ has learned to assign high scores only to anatomically realistic consecutive pairs, minimizing $1 - C(\cdot)$ encourages the predicted masks $\hat{y}_i$ to lie on a manifold of plausible slice-to-slice shapes. This reduces gross mis-localization and discontinuities that often occur in difficult AAA cases.

\subsection{XAI Field and Focus Alignment}

\subsubsection{Gradient-based XAI Field}
\label{sec:xai-field}

To extract a spatial attribution field that reveals the internal focus of the encoder, we attach forward and backward hooks to the final transformer block. Let $\mathcal{A}\in\mathbb{R}^{B\times C\times H_e\times W_e}$ denote the block activations and $\mathcal{G}\in\mathbb{R}^{B\times C\times H_e\times W_e}$ their gradients with respect to a scalar surrogate objective $s_{\mathrm{sur}}$. From these gradients, we compute channel coefficients $w_c$ by spatially averaging each gradient map, i.e., $w_c = (H_eW_e)^{-1}\sum_{i,j}\mathcal{G}_{c,i,j}$. Using these coefficients, we form a coarse attribution map by aggregating activation channels via a positively truncated linear operator,

\begin{equation}
    \Phi_{\mathrm{low}}
= \Omega\!\left(\sum_{c} w_c\,\mathcal{A}_c\right) \in \mathbb{R}^{1 \times H_e \times W_e}
\end{equation}

where $\Omega(\cdot)$ denotes a one-sided gating transformation applied elementwise. Subsequently, this attribution map is lifted to the resolution of the segmentation using a spatial lifting operator $\mathcal{U}$, giving $\Phi = \mathcal{U}(\Phi_{\mathrm{coarse}})\in\mathbb{R}^{1\times H\times W}$. To obtain a probability-like focus distribution, we normalize by its global mass, 

\begin{equation}
    \quad
    \tilde{\phi}_{i,j} = \frac{\phi_{i,j}}{\sum_{u,v} \phi_{u,v} + \epsilon}.
\end{equation}

where $\varepsilon$ ensures numerical stability. The resulting $\tilde{\Phi}$ functions as a dense, attribution-based focus map that summarizes where the encoder concentrates its attention, driven by the current loss.

\subsubsection{Probability Focus Alignment}
We treat both the segmentation probabilities and the XAI field as spatial distributions to enforce agreement between the predictive field of the model and the attribution map.
\begin{equation}
    p = \sigma(s), \quad
    \tilde{p}_{i,j} = \frac{p_{i,j}}{\sum_{u,v} p_{u,v} + \epsilon}.
\end{equation}
The raw probability surface is obtained by applying a nonlinear squashing operator to the logits $p$, and subsequently transformed into a mass–normalized density through $\tilde{p}_{i,j}$. We impose the alignment through two complementary interaction terms. First, a mass-overlap functional compares $p$ with the unnormalized attribution $\Phi$ by evaluating an overlap ratio of their pointwise products.
\begin{equation}
    \mathcal{L}_{\mathrm{ovlp}}
    = 1 - \frac{2 \sum_{i,j} p_{i,j} \phi_{i,j} + \epsilon}{\sum_{i,j} p_{i,j} + \sum_{i,j} \phi_{i,j} + \epsilon}.
\end{equation}

Second, a distributional divergence penalizes discrepancies between the normalized fields $\tilde{p}$ and $\tilde{\Phi}$ by contrasting their local log-masses. The divergence term is the only component maintained in explicit form.

\begin{equation}
    \mathcal{L}_{\mathrm{div}}
    = \mathbb{E}_{i,j}\big[\tilde{p}_{i,j}\log(\tilde{p}_{i,j} + \epsilon) - \tilde{p}_{i,j}\log(\tilde{\phi}_{i,j} + \epsilon)\big]
\end{equation}

By aligning focus mass distributional divergence ($\mathcal{L}_{div}$), the model learns a robust attention manifold that suppresses false distractors. The final alignment objective couples the overlap term and the divergence through a tunable coefficient $\lambda_{\mathrm{div}}$, yielding $\mathcal{L}_{\mathrm{align}}=\mathcal{L}_{\mathrm{ovlp}}+\lambda_{\mathrm{div}}\,\mathcal{J}_{\mathrm{div}}$. This composite penalty encourages the segmentation probability mass to concentrate on regions that the encoder itself identifies as important, thereby discouraging scattered or misaligned focus. Practically, the attribution field $\Phi$ is obtained by differentiating the surrogate scalar $s_{\mathrm{sur}}$ with respect to the encoder representation and applying the above definition of normalization.

\subsection{Anatomy-aware and Confidence Prior}

To further protect thin aneurysm walls and lumen boundaries, we incorporate the learning signal with structure-sensitive penalties derived from differential operators. Let $\mathbb{D}_x$ and $\mathbb{D}_y$ denote first-order image differential operators implemented via discrete convolution kernels, and let $p^\ast=\Psi(s)$ denote the nonlinear squashing of the logits. Applying $(\mathbb{D}_x,\mathbb{D}_y)$ to both $p^\ast$ and $y$ yields two gradient-magnitude fields,
\begin{align}
    G_p=\big((\mathbb{D}_x p^\ast)^2 + (\mathbb{D}_y p^\ast)^2 + \varepsilon\big)^{1/2}\\
    G_y=\big((\mathbb{D}_x y)^2 + (\mathbb{D}_y y)^2 + \varepsilon\big)^{1/2}
\end{align}

whose spatial alignment is measured using a normalized inner-product ratio embedded directly into the penalty term. 

\begin{equation}
    \mathcal{L}_{\mathrm{edge}} =
    1 - \frac{2 \sum_{i,j} G_{p,i,j} G_{y,i,j} + \epsilon}{\sum_{i,j} G_{p,i,j} + \sum_{i,j} G_{y,i,j} + \epsilon}
\end{equation}

Alongside this differential alignment, we impose a topology-sensitive component based on a differentiable medial-axis surrogate. We adopt a differentiable approximation of skeletonization to obtain centerlines $S(p)$ and $S(y)$, and compute
\begin{equation}
    \mathcal{L}_{\mathrm{cline}} =
    1 - \frac{2 \sum_{i,j} S(p)_{i,j} S(y)_{i,j} + \epsilon}{\sum_{i,j} S(p)_{i,j} + \sum_{i,j} S(y)_{i,j} + \epsilon}
\end{equation}

The combination of these differential and topological terms produces an anatomy-regularizing functional,
\begin{equation}
\mathcal{L}_{\mathrm{anat}}
= \tfrac{1}{2}\,\mathcal{L}_{\mathrm{edge}}
+ \tfrac{1}{2}\,\mathcal{L}_{\mathrm{cline}},    
\end{equation}

where each constituent term highlights the normalized correlation structure described above. These components ensure that the model respects both the thickness and the medial topology of the aneurysm region rather than relying solely on regional-level overlap. 

In parallel, the auxiliary branch provides a spatial prior $m_c$ that reflects the model's internal confidence. This prior is trained to match the base probability field through a discrepancy functional $\mathcal{L}_{\mathrm{conf}}$, obtained by applying a monotone contrast operator between $m_c$ and $p^\ast$. The geometric box signal, predicted via the projection head, is penalized through an $\ell_1$ deviation between the predicted and ground-truth bounding vectors. Thus, giving rise to an additional geometric term $\mathcal{L}_{\mathrm{box}}$.

The complete optimization strategy proceeds as a curriculum in which all loss components coexist within a single unified objective, but are emphasized differently across training stages. Let
\begin{equation}
\mathcal{L}_{\mathrm{seg}}=\Lambda_{\mathrm{main}}(s,y)
        +\gamma_{\mathrm{aux}}\,\Lambda_{\mathrm{aux}}(a,y)    
\end{equation}

denote the region-matching component that acts on both primary and auxiliary decoding paths, where each $\Lambda$ subsumes both volumetric and overlap-based discrepancies. The structural regularizer $\mathcal{L}_{\mathrm{anat}}$, the confidence coherence term $\mathcal{L}_{\mathrm{conf}}$, the geometric deviation penalty $\mathcal{L}_{\mathrm{box}}$, and the encoder–attribution alignment functional $\mathcal{L}_{\mathrm{div}}$ together form a pool of auxiliary constraints. During the early phase, optimization prioritizes inter-slice plausibility through a weighted consistency energy $\mathcal{L}_{\mathrm{pair}}$, producing a stage-weighted objective
\begin{equation}
    \mathcal{J}_{\mathrm{early}}
= \mathcal{L}_{\mathrm{seg}}
+ \alpha_1\,\mathcal{L}_{\mathrm{conf}}
+ \alpha_2\,\mathcal{L}_{\mathrm{pair}}
\end{equation}

which stabilizes volumetric coherence and suppresses abrupt spatial discontinuities. Once stable, the emphasis is progressively shifted towards attribution-probability and anatomical faithfulness through
\begin{equation}
\mathcal{J}_{\mathrm{late}}
= \mathcal{L}_{\mathrm{seg}}
+ \beta_1\,\mathcal{L}_{\mathrm{anat}}
+ \beta_2\,\mathcal{L}_{\mathrm{conf}}
+ \beta_3\,\mathcal{L}_{\mathrm{div}}  
\end{equation}

Both stages operate on the same level, and the transition corresponds solely to a shift in emphasis over the same set of losses. This unified formulation ensures that global consistency, boundary geometry, encoder focus alignment, and region-level accuracy are optimized jointly in a compatible and mutually reinforcing manner.

\begin{figure*}[htbp]
    \centering
    \includegraphics[width=\textwidth]{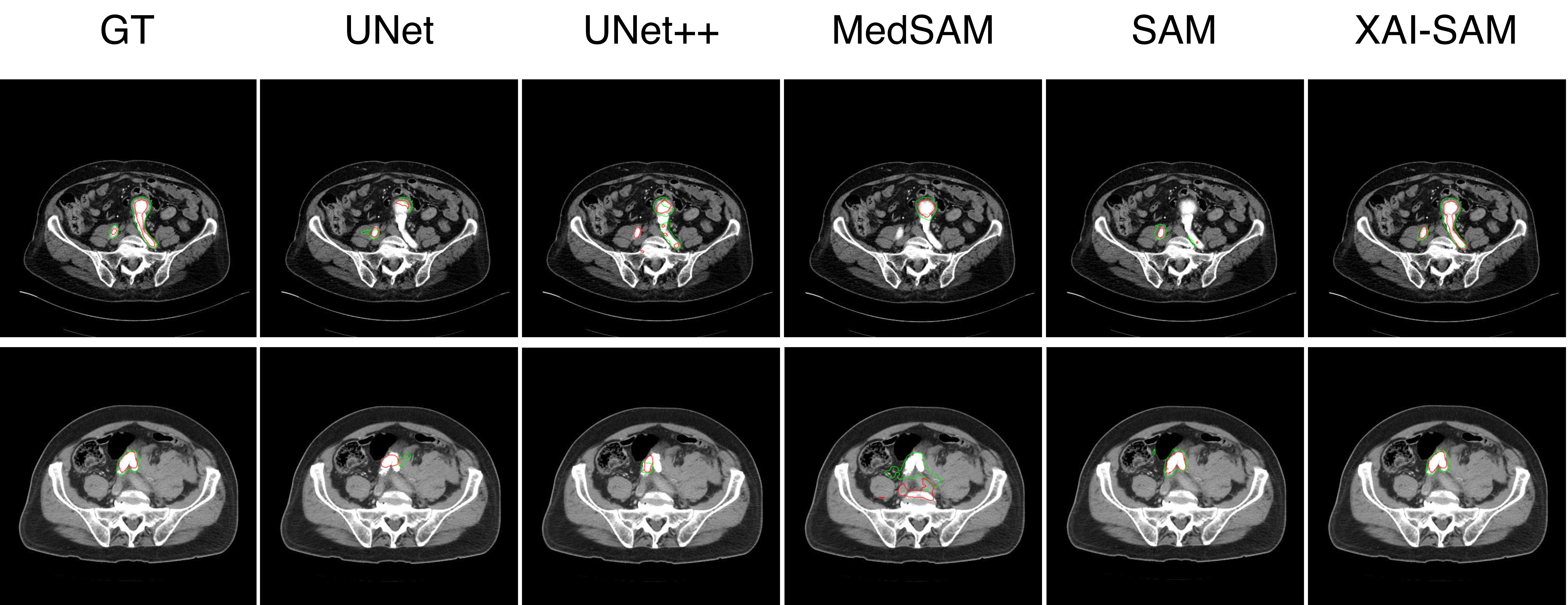}
    \caption{Visualization of model performance on abdominal CT slices. U-Net and U-Net++ frequently mis-localize or leak into adjacent vessels; MedSAM and SAM produce diffuse, unstable masks across slices; XAI-SAM accurately captures thin boundaries and maintains anatomical consistency across the full axial sequence (Red: Lumen. Green: Outer wall).}
    \label{fig:complex-comparison}
\end{figure*}

\section{Experimental Results}

\subsection{Dataset and Evaluation Protocol}
We retrospectively collected 147 contrast-enhanced CTA examinations from Allegheny Health Network and Northwestern Memorial Hospital. All scans had slice thicknesses between 1 and 3 mm and a resolution of 512 x 512. The study was approved by the Institutional Review Boards of the participating institutions. Since this was a retrospective study using de-identified data, informed consent was waived by both IRBs. The data set includes 9,037 axial CTA images used as ground truth, with an 80/20 patient-level split for training and testing.

We evaluated our method on the clinically validated AAA data set that contains two masks per slice: an outer aneurysm wall mask and a lumen mask. The data set was divided into training and test sets, with the test set further divided into two parts. One part is the general test set that contains a wide range of cases, while the second part is a curated complex subset that consists of slices where the baseline models are prone to failure. We report results separately for outer wall and lumen segmentation. For evaluation, we compute the following metrics: Intersection-over-Union (IoU), Dice coefficient, and the 95th percentile Hausdorff distance (HD95). All metrics are reported as mean values across test slices.

% The study was approved by the Institutional Review Boards of the participating institutions at Allegheny General Hospital (Pittsburgh, PA) and Northwestern University’s Feinberg School of Medicine (Chicago, IL).

\subsection{Baselines and Implementation Details}
All models are implemented in PyTorch and trained on eight NVIDIA H200 GPUs for 200 epochs using a batch size of 12 volumes. Training XAI-SAM on 8 NVIDIA H200 GPUs took $\sim$14 hours. At inference, the overhead is negligible ($\sim$12ms per slice) as the gradient hooks are only active during training. We compared against the following baselines: U-Net, U-Net++, nnU-Net, MedSAM, and SAM trained on the AAA data set. The MedSAM fine-tuned model and the SAM-based baseline use the same ViT-B encoder and decoder architecture as our method but without XAI guidance, auxiliary decoder, or pairwise consistency. For the XAI-guided model, we use \( \lambda_{\mathrm{kl}} = 0.2 \) and a gradient-clipping threshold to stabilize training.

\begin{table}[t]
\centering
\caption{General test set results. IoU and Dice are reported in percentages (↑ higher is better); HD95 in millimeters (↓ lower is better). Best results are in \textbf{bold}, second best are \underline{underlined}.}
\label{tab:general}
\small
\begin{tabular}{@{}lccc@{}}
\toprule
Model & IoU (\%)$\uparrow$ & Dice (\%)$\uparrow$ & HD95 (mm)$\downarrow$ \\
\midrule
\multicolumn{4}{c}{\textbf{Outer Wall}}\\
\midrule
U\textendash Net \cite{unet}         & 68.81 & 77.21 & 14.98 \\
U\textendash Net++ \cite{unetpp}       & 68.23 & 78.13 & 22.15 \\
MedSAM \cite{medsam}              & 77.97 & 85.58 &  6.30 \\
nnU-Net \cite{isensee2021nnu} & 80.26 & 86.62 & 6.87 \\
SAM baseline \cite{sam}             & \second{81.53} & \second{88.99} & \second{4.09} \\
\rowcolor{gray!15}
\textbf{XAI\textendash SAM} & \best{96.08} & \best{97.93} & \best{0.19} \\
\midrule
\multicolumn{4}{c}{\textbf{Lumen}}\\
\midrule
U\textendash Net \cite{unet}         & 81.05 & 87.61 & 10.93 \\
U\textendash Net++ \cite{unetpp}       & 71.22 & 81.45 & 30.31 \\
MedSAM \cite{medsam}              & 76.83 & 84.67 &  7.67 \\
nnU-Net \cite{isensee2021nnu} & 80.42 & 87.80 & 12.14 \\
SAM baseline \cite{sam}             & \second{85.83} & \second{91.86} & \second{1.97} \\
\rowcolor{gray!15}
\textbf{XAI\textendash SAM} & \best{95.56} & \best{97.67} & \best{0.26} \\
\bottomrule
\end{tabular}
\end{table}

\begin{table}[t]
\centering
\caption{Complex test set results. IoU and Dice are reported in percentages (↑ higher is better); HD95 in millimeters (↓ lower is better). Best results are in \textbf{bold}, second best are \underline{underlined}.}
\label{tab:complex}
\small
\begin{tabular}{@{}lccc@{}}
\toprule
Model & IoU (\%)$\uparrow$ & Dice (\%)$\uparrow$ & HD95 (mm)$\downarrow$ \\
\midrule
\multicolumn{4}{c}{\textbf{Outer Wall}}\\
\midrule
U\textendash Net \cite{unet}         & 65.14 & 74.15 & 17.98 \\
U\textendash Net++ \cite{unetpp}       & 65.58 & 76.20 & 23.74 \\
MedSAM \cite{medsam}                   & 69.82 & 79.73 & 9.72 \\
nnU-Net \cite{isensee2021nnu} & 76.53 & 84.22 & 10.05 \\
SAM baseline \cite{sam}             & \underline{78.43} & \underline{86.83} & \underline{5.49} \\
\rowcolor{gray!15}
\textbf{XAI\textendash SAM} & \textbf{94.95} & \textbf{97.30} & \textbf{0.31} \\
\midrule
\multicolumn{4}{c}{\textbf{Lumen}}\\
\midrule
U\textendash Net \cite{unet}         & 76.52 & 84.33 & 13.39 \\
U\textendash Net++ \cite{unetpp}       & 66.96 & 78.50 & 31.39 \\
MedSAM \cite{medsam}                   & 70.09 & 79.57 & 10.89 \\
nnU-Net \cite{isensee2021nnu} & 75.51 & 84.17 & 18.07 \\
SAM baseline \cite{sam}             & \underline{82.11} & \underline{89.57} & \underline{2.59} \\
\rowcolor{gray!15}
\textbf{XAI\textendash SAM} & \textbf{94.55} & \textbf{97.11} & \textbf{0.49} \\
\bottomrule
\end{tabular}
\end{table}

\subsection{General Test Set Performance}
In the general test set, as seen in Table \ref{tab:general}, our XAI-SAM substantially outperforms all baselines for both the outer wall and lumen segmentation. For the outer wall, the SAM baseline achieves an IoU of approximately $82\%$ and a Dice of around $89\%$, while XAI-SAM reaches an IoU of approximately $96\%$ and a Dice of $98\%$, with corresponding improvements in HD95. For the lumen, the gains are similarly pronounced: IoU improves from roughly $86\%$ (SAM baseline) to $96\%$ (XAI-SAM), and Dice increases from about $92\%$ to $98\%$. U-Net and U-Net++ show noticeably lower and more variable performance, particularly on outer wall segmentation, where their IoU and Dice have higher standard deviations and much larger HD95 values. The MedSAM fine-tuned model improves over the U-Net variants, but still falls short of the SAM baseline and significantly behind XAI-SAM. Overall, the results of the general test set demonstrate that integrating the alignment of the XAI field, anatomy-aware losses, and confidence priors into the SAM backbone yields a consistent boost across all metrics for both the outer wall and lumen regions.

\subsection{Complex Case Performance}
In Table \ref{tab:complex}, for the complex subset, the benefits of our method become even more apparent. For outer wall segmentation, the SAM baseline achieves an IoU of approximately $78\%$ and a Dice of $87\%$, while XAI-SAM improves these to approximately $95\%$ and $97\%$, respectively. HD95 is reduced by a large margin, reflecting more accurate boundary alignment in difficult cases. Similar trends hold for lumen segmentation, where XAI-SAM improves both overlap measures and reduces boundary errors.

Importantly, U-Net and U-Net++ exhibit a large performance degradation and variance in this subset, with IoU dropping into the $65\%$-$75\%$ range and HD95 values sometimes exceeding $30$-$50$ millimeters (mm). MedSAM performs better but still suffers from over-segmentation and leakage in failure-prone slices. By contrast, XAI-SAM maintains high overlap and low HD95, indicating that the model not only generalizes well but is particularly robust where other models fail.

\subsection{Ablation Study}
To isolate the impact of individual components, we present an ablation study on the complex test set. S: Segmentation Refinement, X: XAI-Field Guidance, A: Anatomy-Aware Loss. Table~\ref{tab:ablation} reports a component-wise ablation in the complex test set. The results confirm that, while Anatomy (A) provides a strong prior, full XAI-SAM integration is required to achieve peak precision and robustness against distractors.

\begin{table}[h]
\centering
\small
\caption{Ablation study on the complex test set.}
\label{tab:ablation}
\begin{tabular}{@{}ccc|ccc|ccc@{}}
\toprule
\multicolumn{3}{c|}{\textbf{Comp.}} & \multicolumn{3}{c|}{\textbf{Outer Wall}} & \multicolumn{3}{c}{\textbf{Lumen}} \\
\textbf{S} & \textbf{X} & \textbf{A} & \textbf{IoU} & \textbf{Dice} & \textbf{HD95} & \textbf{IoU} & \textbf{Dice} & \textbf{HD95} \\ \midrule
$\times$ & $\times$ & $\times$ & 78.4 & 86.8 & 5.4 & 82.1 & 89.6 & 2.6 \\
% $\times$ & $\times$ & $\checkmark$ & 93.8 & 96.7 & 0.5 & 94.4 & 97.0 & 0.4 \\$\times$ & $\checkmark$ & $\times$ & 10.4 & 16.3 & 27.0 & 17.0 & 26.7 & 42.1 \\
% $\times$ & $\checkmark$ & $\checkmark$ & 93.8 & 96.7 & 0.7 & 15.3 & 26.2 & 9.0 \\
$\times$ & $\checkmark$ & $\checkmark$ & 92.9 & 96.2 & 0.4 & 91.3 & 95.4 & 0.6 \\
$\checkmark$ & $\times$ & $\checkmark$ & 93.3 & 96.4 & 0.6 & 90.0 & 94.6 & 1.0 \\
$\checkmark$ & $\checkmark$ & $\times$ & 94.7 & 97.1 & 0.2 & 93.8 & 96.7 & 0.6 \\ \midrule
\rowcolor{gray!15} $\checkmark$ & $\checkmark$ & $\checkmark$ & \textbf{95.0} & \textbf{97.3} & \textbf{0.3} & \textbf{94.6} & \textbf{97.1} & \textbf{0.5} \\ \bottomrule
\end{tabular}
\end{table}

\subsection{Qualitative Case Studies and XAI Analysis}
To better understand why XAI-SAM improves performance, we conducted detailed qualitative case studies on challenging abdominal CT slices, including complex aneurysm morphology, thin and irregular walls, and the presence of adjacent vessels with similar intensity patterns (Fig.~\ref{fig:complex-comparison}). In many cases, U-Net and U-Net++ focus on adjacent vessels or wall-like structures, leading to mis-localization. These confusions are expected, as these architectures rely heavily on local texture cues and lack explicit mechanisms for regulating focus. MedSAM and the standard SAM baseline exhibit a different failure profile. Although they often identify a coarse region of interest, internal attention tends to be diffuse, with large portions of the probability mass spreading into surrounding tissue. This diffuse focus manifests itself in predictions that leak beyond the aneurysm wall or collapse inward near thin boundary segments. In contrast, XAI-SAM’s XAI field is strongly concentrated on the true aneurysm wall and lumen boundaries. The aligned focus is reflected in the predicted masks, which follow the thin wall more closely and avoid leakage. For volumes in which pairwise consistency was applied, we observe that the slice-to-slice masks remain more stable and anatomically coherent. Abrupt changes in segmentation, such as sudden disappearance or expansion of the aneurysm region, are substantially reduced compared to baselines. Together, these qualitative findings support our claim that treating encoder focus as a training signal helps correct specific classes of model failures that are not fully addressed by standard region-based losses.

\section{Discussion}

Our experiments show that explicitly optimizing the encoder’s internal focus via an XAI field and integrating this signal into both representation and decoding leads to more reliable AAA segmentation, especially in complex cases. The improvements are not limited to global metrics; they also manifest as better preservation of thin aneurysm walls, reduced over-segmentation, and more consistent behavior across slices. From a methodological standpoint, our approach bridges post-hoc explainability and training-time supervision. Instead of using attribution solely to analyze models after training, we use it proactively to shape the learned representation. The XAI alignment loss encourages the model to look where it predicts, and anatomy-aware losses ensure that this focus remains faithful to clinically meaningful structures. The pairwise consistency classifier complements this by enforcing a learned notion of slice-to-slice plausibility, helping to suppress anatomically implausible transitions. There are several limitations and avenues for future work. First, our XAI field is based on attribution, which may not capture all aspects of the encoder’s decision process. Exploring alternative attribution methods or learned attention maps as the basis for the XAI field is an interesting direction. Second, we currently train separate models for outer wall and lumen masks; a multi-task formulation that jointly models both regions and their relationships could further improve performance. Third, while our experiments focus on AAA, the proposed framework is generic and could be applied to other vascular or organ segmentation tasks where model failures are tied to misallocated attention. In summary, our results suggest that explainability-driven analysis can be elevated from a diagnostic tool to a guiding principle for training segmentation models. By dissecting how and where models fail and then explicitly encoding this knowledge into the learning objective, we can obtain segmentation systems that are not only more accurate but also more aligned with clinical intuition in the scenarios that matter most.

\section*{Acknowledgement}
This work was funded by the National Institutes of Health (Grant No. R01HL159300).

{
    \small
    \bibliographystyle{ieeenat_fullname}
    \bibliography{main}
}


\end{document}